# Deep learning based detection of collateral circulation in coronary angiographies

Cosmin-Andrei Hatfaludi [1, 2], Daniel Bunescu [1, 2], Costin Florian Ciuşdel [1, 2], Alex Serban [1, 2], Karl Böse [3], Marc Oppel [3], Stephanie Schröder [3], Christopher Seehase [3], Harald F. Langer, M.D. [3, 4, 6, 7], Jeanette Erdmann [4, 7], Henry Nording, M.D. [3, 4, 5], Lucian Mihai Itu [1, 2]

1 Advanta, Siemens SRL, 15 Noiembrie Bvd, 500097, Brasov, Romania.

2 Automation and Information Technology, Transilvania University of Brasov, Mihai Viteazu nr. 5, 5000174, Brasov, Romania.

3 Cardioimmunology group, Medical Clinic II, University Heart Center Lübeck, 23538 Lübeck, Germany.

4 DZHK (German Centre for Cardiovascular Research), partner site Hamburg/Lübeck/Kiel, 23562 Lübeck, Germany.

5 University Hospital, Medical Clinic II, University Heart Center Lübeck, 23538 Lübeck, Germany.

6 Department of Cardiology, Angiology, Haemostaseology and Medical Intensive Care, University Medical Centre Mannheim, Medical Faculty Mannheim, Heidelberg University, Mannheim, Germany.

7 German Centre for Cardiovascular Research (DZHK), Partner Site Heidelberg/Mannheim, Germany

*Abstract*— Coronary artery disease (CAD) is the dominant cause of death and hospitalization across the globe. Atherosclerosis, an inflammatory condition that gradually narrows arteries and has potentially fatal effects, is the most frequent cause of CAD. Nonetheless, the circulation regularly adapts in the presence of atherosclerosis, through the formation of collateral arteries, resulting in significant long-term health benefits. Therefore, timely detection of coronary collateral circulation (CCC) is crucial for CAD personalized medicine. We propose a novel deep learning based method to detect CCC in angiographic images. Our method relies on a convolutional backbone to extract spatial features from each frame of an angiography sequence. The features are then concatenated, and subsequently processed by another convolutional layer that processes embeddings temporally. Due to scarcity of data, we also experiment with pretraining the backbone on coronary artery segmentation, which improves the results consistently. Moreover, we experiment with few-shot learning to further improve performance, given our low data regime. We present our results together with subgroup analyses based on Rentrop grading, collateral flow, and collateral grading, which provide valuable insights into model performance. Overall, the proposed method shows promising results in detecting CCC, and can be further extended to perform landmark based CCC detection and CCC quantification.

## I. INTRODUCTION

Cardiovascular disease (CVD) is the leading cause of mortality and morbidity for the entire world population [1]. The most frequent cause of CVD is atherosclerosis [2], an inflammatory disease that gradually obstructs arteries and has life-threatening effects when present in the coronary circulation, i.e. leading to coronary artery disease (CAD). While traditional CAD risk factors (e.g., age, gender, diabetes, etc.) [3], as well as genetical risk factors [4] are highly predictive of the onset of CVD, mortality and morbidity is determined by clinical events such as the occurrence of ischemic tissue damage, which cannot be well predicted from traditional CAD risk factors [5-7].

One of the factors which helps to avoid ischemic tissue damage, is the development of coronary collateral circulation (CCC) [8]. CCC is characterized by natural bypasses (collateral arteries) which start to form as a result of progressive blood vessel lumen constriction and blood flow restrictions. Recent clinical studies have demonstrated that well-functioning CCC can independently predict lowered mortality and improved survival rates [9,10]. The protective effect translates to improved left ventricular (LV) function, decreased remodelling, and a lower risk of life-threatening arrhythmias [11]. Thus, CCC can be regarded as a survival advantage [12].

However, technical methods to automatically detect CCC currently lack behind studies on the clinical benefits of CCC. We aim to bridge this gap by introducing a novel method to detect the presence of CCC using deep neural networks on invasive coronary angiography (ICA) images. CCC detection represents an important prerequisite for assessing the morbidity rate for patients with CAD. Moreover, ICA is the gold standard in CAD. To the best of our knowledge, this is the first study to attempt CCC detection on ICA.

The remainder of this article is organized as follows. We start by discussing background information and related work (Section II). Next, we introduce the available data sets and the model architecture (Section III), followed by empirical results (Section IV), a discussion (Section V) and conclusions (Section VI).

## II. BACKGROUND AND RELATED WORK

As mentioned earlier, no previous studies have attempted to detect CCC from ICA. Nevertheless, some studies focused on the assessment of CTO (chronic total occlusion) collaterals from ICA, detecting the onset of coronary artery disease, and on the evaluation of collaterals in patients with ischemic stroke using CT brain scans.

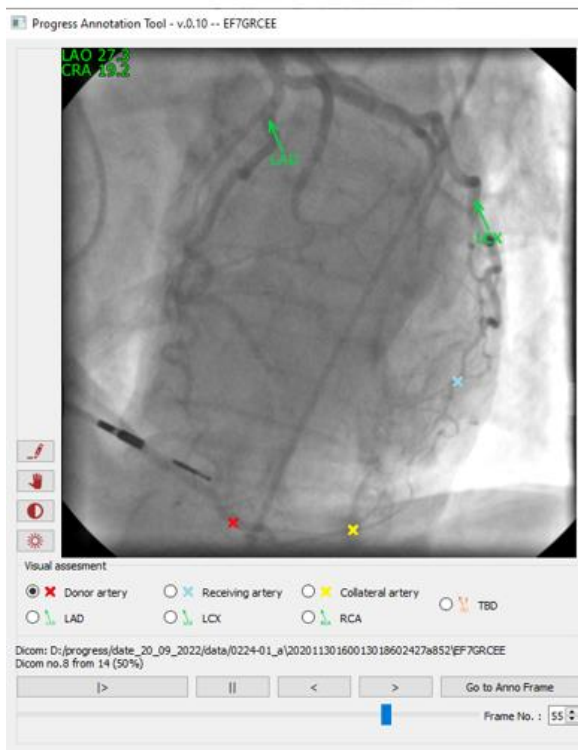

Fig. 1. Python based annotation tool allowing the clinical experts to perform the annotation of ICAs with CCC.

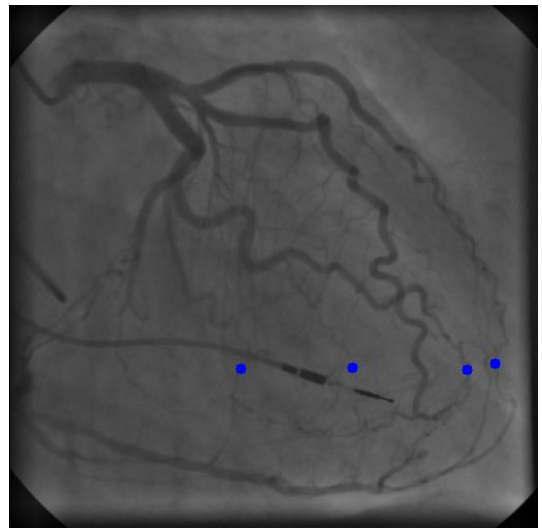

Fig 2. Sample annotation for CCC.

L. Liu et al. [13] proposed a deep learning based method for collateral physiology assessment under total occlusion conditions. Their model automatically extracts the length or time curves of the coronary filling to perform the assessment. H. Kuang et al. [14] proposed a vessel segmentation method for automating collateral scoring on brain CT angiography using a hybrid CNN transformer network. L. Wolff et al. [15] assessed a commercial algorithm for automated collateral scoring on brain CT angiography. Unfortunately, technical descriptions of the algorithm are not provided. Nevertheless, when tested against experts, the agreement between the algorithm and the experts is not significant.

M. Aktar et al. [16] proposed a deep learning method for collateral evaluation in patients with ischemic stroke using CT angiography. Their model processes 2D slices from CTA and uses a voting scheme to determine the outcomes. An additional difficulty of CCC detection is that, due to the myocardial contraction, the coronary vessels display a significant motion during a cardiac cycle.

Moreover, processing ICAs or CTAs with deep learning methods has a long background, for tasks such as coronary arteries segmentation [17], calcium scoring [18] or cardiovascular hemodynamic prediction on bypass surgeries [19].

Since the available data set for our task is small, a relevant related field is that of few-shot learning, with applications to deep learning. FSL is a learning method in which predictions are performed based on a limited number of samples [20].

Few shot learning (FSL) methods are classified into four categories: multitask learning, embedding learning, learning with external memory, and generative modeling. Herein we use prototypical networks [21], a subcategory of the embedding learning models, considered the state-of-the-art for classification tasks.

III. DATA AND CCC DETECTION

*A. Dataset*

We operate in a low data regime, using a data set that consists of only 88 patients (with 168 ICAs) for which CCC was annotated, and 90 patients (with 168 ICAs) without CCC. Some patients have multiple ICAs but for each patient we have at least one ICA. For the ICAs with CCC, only one frame was annotated, marking the location of the CCC, as illustrated in Fig. 1.

To annotate the data, we developed a custom, Python-based, annotation tool that allows clinical experts to perform the annotation of ICAs with CCC (see Fig 1.). The tool enables experts to annotate the location of the collateral artery, the donor artery and the receiving artery, Rentrop grading [22], pathways [23], collateral flow grade [24], blush grade [25], the donor segment, the receiving segment, and the collateral artery size. We use these annotations for the subgroup analysis and for a future CCC quantification model. To reduce annotation time, the experts were asked to annotate only one frame for each angiography with CCC (as illustrated in Fig. 2)

The annotators were trained and supervised by expert interventional cardiologists with at least 10 years of experience in the catheter lab, and annotations were checked reciprocally and, in case of disagreement, discussed by at least two independent experts.

*B. CCC detection*

To perform CCC detection while also balancing hardware resources, we extract from each angiography 11 consecutive frames. For the sequences annotated with CCC, we use five frames before the annotated frame, the annotated frame, and five frames after the annotated frame.

While more frames may be used, the computational costs increase significantly, and the information gains decrease since the contrast flushes out over time.

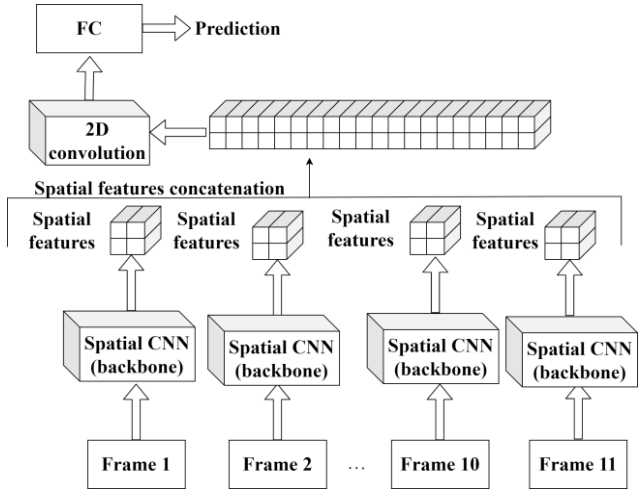

Fig 4. The proposed method for the CCC detection.

For the sequences without CCC, we employ a model that is described in section III.C to compute the vesselness score and select 11 frames centered around the frame with the highest vesselness score. All data is normalized using z-score normalization [26] across the spatial pixel intensities.

To extract spatial features from the selected frames we use a spatial CNN (backbone) with six layers meant to capture a large field of view. The extracted spatial features are later concatenated and processed by 2D convolution with a kernel size of one, that extracts spatial and temporal features from the sequence. Based on the extracted characteristics, a FC layer outputs a binary decision for each frame (whether CCC is present or not). For this reason, a sigmoid function [27] has been applied at the output of the network. If the predicted value is greater than 0.5, the CCC is present, and if the predicted value is less than 0.5 the CCC is not present. The goal of this layer is to extract both spatial and temporal features from all frames. An illustration of our approach is depicted in Fig. 4.

The spatial CNN is used in three configurations: no pretraining, pretrained without freezing the weights of the backbone, and pretrained with freezing the weights (and training just the last convolutional and fully connected layers). The pretrained backbone uses an auxiliary task of segmenting the vessels, as discussed in Section III.B.

The models are trained using a vanilla method and a method based on prototypical networks. Due to the low data regime we operate in, we trained the models using k-fold cross validation [28] with k = 4, for 100 epochs. The k-fold cross validation datasets are split at patient level, meaning that all ICAs from one patient belong to the same fold. The classification accuracy is computed for each epoch, and the epoch leading to the highest accuracy on the entire dataset (all folds) is chosen for reporting the statistics. All models are trained using Adam optimizer [29] and a fixed learning rate of 0.0001.

### C. Backbone pretraining

To boost the performance of the vanilla model, we also pretrained the backbone on a proxy task and performed transfer learning on the CCC task.

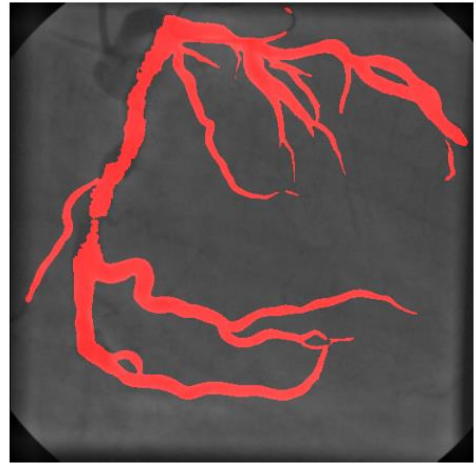

Fig 3. The predicted vesselness segmentation for a dicom from our dataset.

For pretraining, we used a supervised learning vessel segmentation task, for which annotations were already available. An illustration of the task is provided in Fig 3.

To generate ground truth masks, we had access to annotated artery centerlines and diameter information. From these, we generated segmentation masks using a Gaussian function (with mean centered on the centerline points and a standard deviation of 0.75). The Gaussian ensured smooth vessel edges. To preserve smoothness, we formulated the training process as a regression problem, and used MSE as loss (instead of thresholding the segmentation masks and employing classification loss functions). In total, we had access to 3350 ICAs for pretraining, from which we used 70% for training, 10% for validation and 20% for testing. Training ran for 100 epochs and we used early stopping based on the minimum loss on the validation data set. The evaluation on the test set revealed a Dice score of 0.92 (with sensitivity 0.95 and specificity 0.94).

## IV. EXPERIMENTAL RESULTS

### A. CCC detection

To evaluate the CCC detection performance, we determined the accuracy (Acc.), sensitivity (Sens.) and specificity (Spe.) [30]. The results obtained on the entire dataset following cross-validation, using all models described in Section III, are illustrated in Table I. The table indicates that the best results are obtained using a FSL model with pretraining and with weight freeze. All models perform better if the backbone is pretrained. Fig. 5 displays four sample cases from the dataset: one true positive (TP), one true negative (TN), one false positive (FP), and one false negative (FN). We observe that both the FP and the FN samples are difficult, as the FP image has artifacts which can easily be confused with CCC and the FN sample's field of view looks incomplete.

Table I. Results obtained for the CCC detection.

| Model | Pretrain | Freeze | Acc. [%] | Sens. [%] | Spec. [%] |
|---|---|---|---|---|---|
| Classic | ✗ | ✗ | 65.2 | 62.5 | 67.9 |
| Classic | ✓ | ✗ | 78.9 | 79.8 | 78.0 |
| Classic | ✓ | ✓ | 76.2 | 75 | 77.4 |
| FSL | ✗ | ✗ | 61.3 | 61.9 | 60.7 |
| FSL | ✓ | ✗ | 77.7 | 77.4 | 78.0 |
| FSL | ✓ | ✓ | **79.5** | **80.4** | **78.6** |

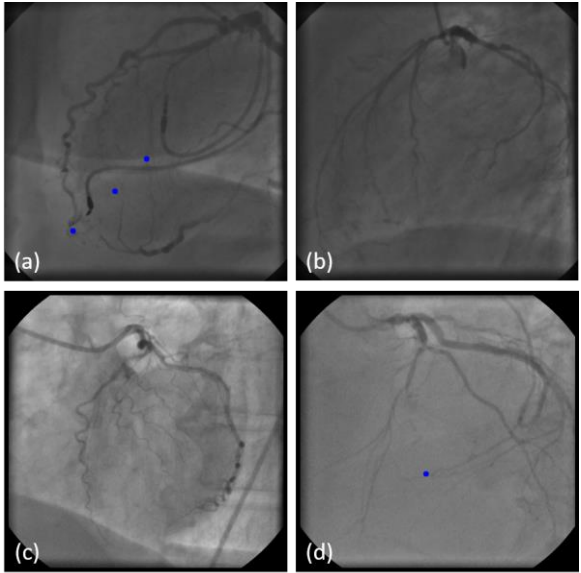

Fig. 5. Four sample cases: one for each of the categories: (a) TP, (b) FP, (c) TN, and (d) FN.

The confusion matrix for the best performing model is depicted in Fig. 6: sensitivity and specificity are balanced.

In the following, we use the best performing model for a series of subgroup analyses. For these analyses, we focus on the dataset of ICAs with CCC. Since only TPs and FNs can be obtained within this dataset, we analyse the sensitivity. The subgroups were defined using the annotations performed by the clinical experts. The first analysis is based on the Rentrop grading (see Table II).

Table II. Results obtained for subgroup analyses based on Rentrop grading.

| Rentrop Grading | Nr. Of samples | Sensitivity |
|---|---|---|
| 1 | 20 | 65 |
| 2 | 51 | 88.23 |
| 3 | 91 | 80.21 |

Rentrop grading one means filling of side branches of the artery to be dilated via collateral channels without visualization of the epicardial segment, Rentrop grading two means partial filling of the epicardial segment via collateral channels, and Rentrop grading three means complete filling of the epicardial segment of the artery dilated via collateral channels. The model detects the collateral circulation Rentrop grading two and three better than the collateral circulation graded with one. This is to be expected, since epicardial segments cannot be visualized well in Rentrop grading one.

Table III. Results obtained for subgroup analyses based on collateral flow grade.

| Collateral flow grade | Nr. Of samples | Sensitivity |
|---|---|---|
| 1 | 10 | 80 |
| 2 | 33 | 75.75 |
| 3 | 46 | 82.6 |
| 4 | 73 | 82.19 |

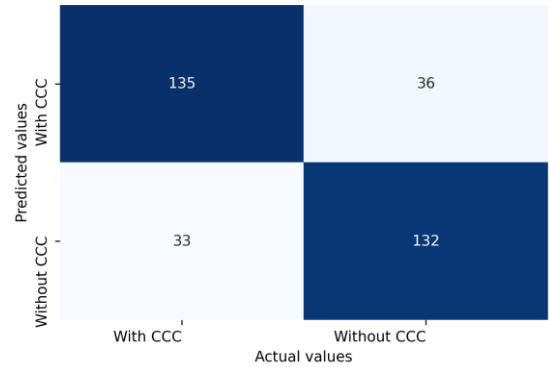

Fig 6. The confusion matrix for the best performing model.

The second analysis is based on the collateral flow grade, which can vary between one and four. Collateral flow grade one means that collateral flow is barely apparent, dye is not visible throughout the cardiac cycle, but in at least three consecutive frames. Collateral flow grade two means that the collateral flow is moderately opaque, the dye being present in more than 75% of cycle. Collateral flow grade three means that the collateral flow is well opacified, with clear antegrade dye motion. Collateral flow grade four means that the collateral flow is well opacified, fills antegrade, with very large vessels (> 0.7 mm). The results obtained for this subgroup analysis are depicted in Table III. The model performs well on all four subgroups, only the results for collateral flow grade two are slightly lower. This indicates that even when the dye is visible on only a few frames, the model can detect the CCC.

The third subgroup analysis focuses on the collateral artery size measured in pixels (most of the ICAs in our dataset do not have the pixel spacing information; hence, it is not possible to perform this analysis in physical units). The collateral artery size was measured using the annotation tool, and three equally sized subgroups have been defined (see Table IV). We observe that the model performs worst on the subgroup with the smallest collateral artery size, which is to be expected since these are the most difficult ones to detect.

Table IV. Results obtained for subgroup analyses based on collateral artery size.

| Collateral artery size | Nr. Of samples | Sensitivity |
|---|---|---|
| < 3.14 | 54 | 74.07 |
| 3.14 – 4 | 54 | 87.03 |
| > 4 | 55 | 80 |

## V. DISCUSSIONS

Overall, the models introduced show promising results for CCC detection, albeit the performance can be further improved. One of the main challenges was the small dataset size. We approached this problem from two angles: (i) by pretraining the models, and (ii) by employing few shot deep learning techniques. However, the lack of data and population diversity induces small inconsistencies in the results, which we discuss in this section.

As shown in Table I, pretraining provided significant improvements for both classical and few shot learning. This motivates further research into better pretraining methods, using more data. For our task, we had annotated data and formulated pretraining and a supervision task. However, if more data without annotations are available, self-supervised methods deserve exploration [31].

The use of few shot learning techniques further provided marginal improvements. However, as shown in Table I, these improvements were not entirely consistent. For example, when using the pretrained backbone without freezing the weights, classical training exceeded few shot learning. This result is likely because optimizing the prototypes and fine-tuning the backbone weights was not possible given limited training data set. When freezing the weights, however, the prototypes could be better defined and the few shot method exceeded classical training. This balance between choosing which parameters to specialize in limited data regimes deserves future exploration.

The subgroup analyses in Section IV also revealed minor uneven results. For example, the sensitivity for Rentrop grading in Table II shows better performance for Rentrop grading two than three. Nevertheless, the number of data points for grading three is significantly higher than for grading two, which induces more diversity in the analyzed population. We expect more even results once more data is available.

The same result can be observed for collateral flow grade sensitivity in Table III, where the sensitivity for collateral flow 1 is higher than for collateral flow 2. These small inconsistencies will likely be removed by analyzing data that is more diverse.

Moreover, the artery size measurements in Table IV revealed that sensitivity is higher for medium artery sizes than for large sizes. This result is likely due to some inherent bias stemming from the pretraining method, where the model was pretrained to avoid segmenting larger structures such as the catheter. Since the best model uses a backbone with frozen weights, it is likely that the model is cautious with larger structures.

## VI. CONCLUSIONS

We proposed a deep learning based method to detect the CCC on ICAs in low data regimes. The method integrates both spatial and temporal features using a CNN-based backbone to extract spatial information from multiple frames, and an additional CNN-based head which integrates temporal features. Due to limited data availability, we experiment both with pretraining the backbone on coronary segmentation and with few-shot learning techniques.

We performed multiple subgroup analyses to understand the outcomes and report on their results. Overall, the best performing model uses both pretraining and few-shot learning, and demonstrates promising results in CCC detection. Future work will focus collecting more data and extending pretraining to self-supervision. Moreover, future work will also focus on extending the current model to perform landmark based CCC detection and CCC quantification, which in turn can benefit the core task of CCC detection.

## ACKNOWLEDGMENT

This work was supported by a grant of the Romanian National Authority for Scientific Research and Innovation, CCCDI – UEFISCDI, project number ERANET-PERMED-PROGRESS, within PNCDI III. This work has been partly funded under the European Union's Horizon 2020 research and innovation programme under grant agreement N° 101017578 (SIMCor: In-Silico testing and validation of Cardiovascular IMplantable devices). The research leading to these results has received funding from the EEA Grants 592 2014-2021, under Project contract no. 33/2021.